\documentclass[12pt,a4paper]{article}

\usepackage{graphicx}
\usepackage[pdftitle={Quant}]{hyperref}
\usepackage{mathtools}
\usepackage{natbib}
\usepackage{url}

\newcommand{\quant}{\textsc{Quant}}
\newcommand{\hydra}{\textsc{Hydra}}
\newcommand{\rocket}{\textsc{Rocket}}
\newcommand{\minirocket}{\textsc{MiniRocket}}
\newcommand{\multirocket}{\textsc{MultiRocket}}

\begin{document}

\title{{\quant} \\ {\large A Minimalist Interval Method for Time Series Classification}}

\author{%
    Angus Dempster, Daniel F. Schmidt, Geoffrey I. Webb\\
    {\normalsize Monash University, Melbourne, Australia}\\
    {\footnotesize \texttt{\{angus.dempster,daniel.schmidt,geoff.webb\}@monash.edu}}%
}

\date{}

\maketitle\vspace*{-2em}%

\begin{abstract}
    We show that it is possible to achieve the same accuracy, on average, as the most accurate existing interval methods for time series classification on a standard set of benchmark datasets using a single type of feature~(quantiles), fixed intervals, and an `off the shelf' classifier.  This distillation of interval-based approaches represents a fast and accurate method for time series classification, achieving state-of-the-art accuracy on the expanded set of 142 datasets in the UCR archive with a total compute time (training and inference) of less than 15~minutes using a single CPU core.
\end{abstract}

\section{Introduction} \label{section-introduction}

\begin{figure}
    \centering
    \includegraphics[width=\linewidth]{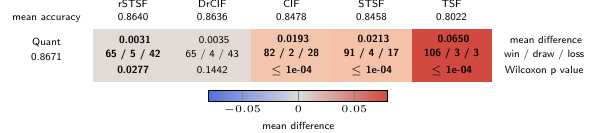}%
    \vspace{-1em}%
    \caption{Multiple Comparison Matrix for {\quant} vs TSF, STSF, rSTSF, CIF, and DrCIF, for a subset of 112 datasets from the UCR archive.}%
    \label{fig-mcm-ucr112}%
\end{figure}

Interval methods represent a long-standing and prominent approach to time series classification.  Most interval methods are strikingly similar, closely following a paradigm established by \citet{rodriguez_etal_2000} and \citet{geurts_2001}, and involve computing various descriptive statistics and other miscellaneous features over multiple subseries of an input time series, and/or some transformation of an input time series (e.g., the first difference or discrete Fourier transform), and using those features to train a classifier, typically an ensemble of decision trees \citep[e.g.,][]{deng_etal_2013,lines_etal_2018}.  This represents an appealingly simple approach to time series classification \citep[see][]{middlehurst_and_bagnall_2022,henderson_etal_2023}.

We observe that it is possible to achieve the same accuracy, on average, as the most accurate existing interval methods simply by sorting the values in each interval and using the sorted values as features or, in order to reduce the size of the feature space (and, accordingly, computational cost), to subsample these sorted values, i.e., to use the quantiles of the values in the intervals as features.  We name this approach {\quant}.

The difference in mean accuracy and the pairwise win/draw/loss between {\quant} and several other prominent interval methods, namely, TSF \citep{deng_etal_2013}, STSF \citep{cabello_etal_2020}, rSTSF \citep{cabello_etal_2021}, CIF \citep{middlehurst_etal_2020}, and DrCIF \citep{middlehurst_etal_2021b}, for a subset of 112 datasets from the UCR archive (for which published results are available for all methods), are shown in the Multiple Comparison Matrix (MCM) in Figure \ref{fig-mcm-ucr112} \citep[see][]{ismailfawaz_etal_2023}.  Results for the other methods are taken from \citet{middlehurst_etal_2023}.  As shown in Figure \ref{fig-mcm-ucr112}, {\quant} achieves higher accuracy on more datasets, and higher mean accuracy, than existing interval methods.  Total compute time for {\quant} is significantly less than that of even the fastest of these methods (see further below).

\begin{figure}[t]
    \centering
    \includegraphics[width=0.65\linewidth]{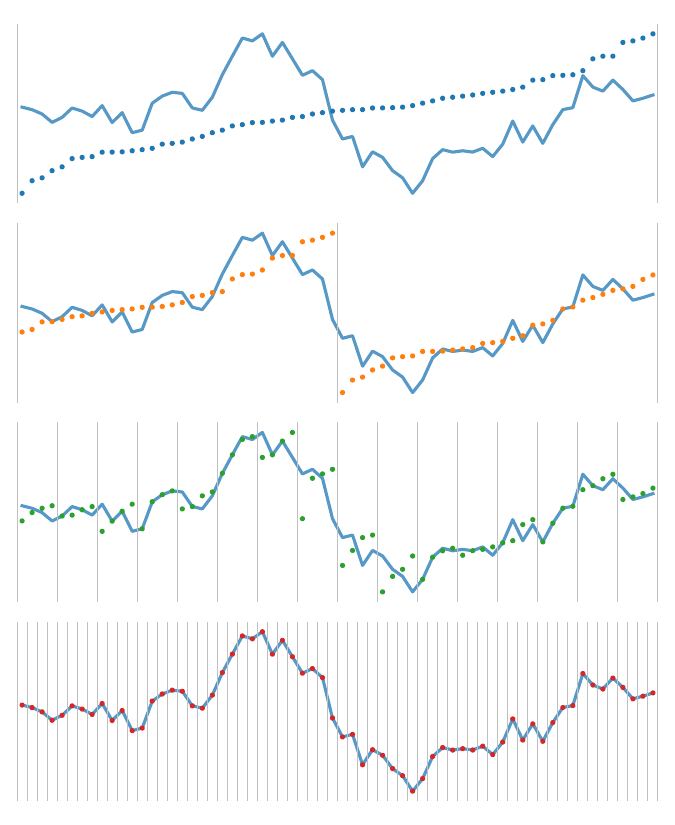}%
    \vspace{-1em}%
    \caption{Sorted values for intervals of decreasing length.  Larger intervals contain more distributional information but less location information.}%
    \label{fig-intervals}%
    \vspace{1em}%
\end{figure}

When using quantiles (or sorted values) as features, as we increase or decrease interval length, we move between two extremes: (a) a single interval where the quantiles (or sorted values) represent the distribution of the values over the whole time series (distributional information without location information); and (b) intervals of length one, together consisting of all of the values in the time series in their original order (location information without distributional information): see Figure \ref{fig-intervals}.

Quantiles represent a superset of many of the features used in existing interval methods (min, max, median, etc.).  Using quantiles allows us to trivially increase or decrease the number of features, by increasing or decreasing the number of quantiles per interval which, in turn, allows us to balance accuracy and computational cost.  We find that quantiles can be used with fixed (nonrandom) intervals, without any explicit interval or feature selection process, and with an `off the shelf' classifier, in particular, extremely randomised trees \citep{geurts_etal_2006}, following \citet{cabello_etal_2021}.

The key advantages of distilling interval methods down to these essential components are simplicity and computational efficiency.  {\quant} represents one of the fastest methods for time series classification.  The cost of computing the quantiles, in particular, is very low.  Median transform time over 142 datasets in the UCR archive is less than one second.  Total compute time (training and inference) is under 15 minutes for the same 142 datasets using a single CPU core.  This is approximately $5 \times$ faster than the fastest existing interval method, rSTSF \citep{cabello_etal_2021}, already one of the fastest methods for time series classification.

The rest of this paper is structured as follows.  In Section \ref{section-background}, we discuss relevant related work.  In Section \ref{section-method}, we set out the key aspects of the method.  In Section \ref{section-experiments}, we present experimental results including a sensitivity analysis.

\section{Background} \label{section-background}

\subsection{Interval Methods}

Methods closely resembling current state-of-the-art interval methods have been applied to the domain of time series classification at least since \citet{rodriguez_etal_2000} and \citet{geurts_2001}.  Most interval methods are strikingly similar, closely following the basic concept set out in, e.g., \citet{rodriguez_and_alonso_2004}, namely:

\begin{itemize}
    \item for a set of intervals (subseries) taken from the input time series, and/or some transformation of the input time series such as the first difference or discrete Fourier transfom;
    \item compute descriptive statistics (e.g., mean and variance) and other features for the values in each interval; and
    \item use the computed features to train a classifier, typically an ensemble of decision trees.
\end{itemize}

Different interval methods are characterised by the set of transformations applied to the input time series, the characteristics of the intervals, the use of interval and/or feature selection, and the choice of classifier.

Many methods use one or more transformations of the input time series.  RISE replaces the input time series with spectral, autocorrelation, and autoregressive representations \citep{lines_etal_2018,flynn_etal_2019}.  More recently, the use of the original input time series in combination with the first difference, and some form of frequency domain representation (e.g., the discrete Fourier transform), has lead to significant improvements in accuracy over earlier methods \citep{cabello_etal_2020,cabello_etal_2021,middlehurst_etal_2021b}.

Some methods use fixed intervals, recursively splitting the input time series in half \citep[e.g.,][]{rodriguez_and_alonso_2004}, while some use random intervals, recursively splitting the input time series at random points, or sampling intervals with random length and position \citep[e.g.,][]{deng_etal_2013,baydogan_etal_2013,cabello_etal_2021}.  Others methods use heuristic approaches \citep[e.g.,][]{cabello_etal_2020,altay_and_baydogan_2021}.

All or almost all proposed methods use a fixed set of descriptive statistics (e.g., mean and variance), often combined with other features such as slope \citep[e.g.,][]{deng_etal_2013,middlehurst_etal_2020}.  Several methods employ some form of explicit feature and/or interval selection process \citep[e.g.,][]{cabello_etal_2020,cabello_etal_2021,li_etal_2023}.

Other variations to the basic concept of interval methods include forming `bag of words' representations of the features extracted from intervals \citep[e.g.,][]{baydogan_etal_2013,baydogan_and_runger_2016}, fitting Gaussian process models to intervals \citep{berns_etal_2021}, and approaches incorporating clustering \citep{schmidt_and_lohweg_2021}.

Most methods use an ensemble of decision trees, including specialised decision trees for interval features such as `time series trees' \citep[e.g.,][]{deng_etal_2013,middlehurst_etal_2020,middlehurst_etal_2021b}, or standard ensembles such as boosted decision trees \citep[e.g.,][]{rodriguez_etal_2001,geurts_2001}, random forests, or extremely randomised trees \citep[e.g.,][]{cabello_etal_2021}.  Some methods use other classifiers such as support vector machines \citep[e.g.,][]{rodriguez_and_alonso_2005}.  In this context, it is worth noting that some earlier methods were proposed prior to the introduction of what are now considered canonical classifiers such as random forests or extremely randomised trees, and prior to or only shortly after the introduction of the UCR archive.

The two most accurate current interval methods on the datasets in the UCR archive are DrCIF \citep{middlehurst_etal_2021b}, and rSTSF \citep{cabello_etal_2021}.  Both, in turn, build on TSF \citep{deng_etal_2013}.  TSF uses random intervals (intervals with random position and length), and computes the mean, variance, and slope of the values in each interval.  TSF uses an ensemble of specialised decision trees (`time series trees'), using a splitting criteria that combines entropy and a tie-breaking procedure, and trains each tree separately using a different set of random of intervals \citep{deng_etal_2013}.  \citet{bagnall_etal_2017} found that TSF was faster and at least as accurate as other interval methods on the datasets in the UCR archive at the time.

DrCIF builds on CIF \citep{middlehurst_etal_2020}, sampling random intervals (random position and length) from the input time series, first difference, and a periodogram, and computes features including the mean, standard deviation, slope, median, interquartile range, min, max, as well as the catch22 features \citep{lubba_etal_2019}.  DrCIF uses a version of `time series trees' as per TSF, training each tree separately with a random set of intervals and a random subset of features.  DrCIF is one of the four components of HIVE-COTE 2 (HC2), the most accurate method for time series classification on the datasets in the UCR archive \citep{middlehurst_etal_2021b}.

rSTSF builds on STSF \citep{cabello_etal_2020}.  For each of the original time series, first difference, a periodogram, and an autoregressive representation (the coefficients of an autoregressive model), and for each of the mean, standard deviation, slope, min, max, median, interquartile range, and two additional features (the number of intersections with the mean and the number of values greater than the mean), rSTSF recursively splits the input at random points, selecting intervals using the Fisher score, performing a kind of interval or feature selection.  Unlike DrCIF, rSTSF uses an `off the shelf' classifier, namely, extremely randomised trees.  While DrCIF and rSTSF produce similar accuracies on the datasets in the UCR archive, rSTSF is considerably faster \citep{middlehurst_etal_2023}.

\subsection{Other State-of-the-Art Methods}

In the recent `bake off redux', \citet{middlehurst_etal_2023} evaluate the most accurate current methods for time series classification over an expanded set of 142 datasets from the UCR archive.  \citet{middlehurst_etal_2023} determine that the most accurate methods from each of a diverse set of different approaches to time series classification are: Proximity Forest, FreshPRINCE, rSTSF, WEASEL-D, InceptionTime, RDST, {\multirocket}+{\hydra}, and HC2.

Proximity Forest (PF) is an ensemble of decision trees using distance measures as splitting criteria \citep{lucas_etal_2019}.  Proximity Forest 2.0 (PF2) is a recent extension of PF that improves computational efficiency, and uses a different set of distance measures \citep{herrmann_etal_2023}.  (PF2 was published concurrently with \citet{middlehurst_etal_2023}, and is not included in the study.)

FreshPRINCE focuses on simplicity, and combines features drawn from the TSFresh feature set, computed over the whole input time series, with a rotation forest classifier \citep{middlehurst_and_bagnall_2022}.

WEASEL-D is a dictionary method, involving extracting and counting symbolic patterns in time series, building on WEASEL \citep{schafer_and_leser_2016}, and uses dilated sliding windows and the Symbolic Fourier Transform with random parameters to extract patterns, in conjunction with a ridge regression classifier \citep{schafer_and_leser_2023}.

InceptionTime is an ensemble of convolutional neural network models based on the Inception architecture \citep{ismailfawaz_etal_2020}, and represents the most accurate deep learning model on the datasets in the UCR archive.

RDST is a shapelet method, computing features based on the distance between input time series and a set of discriminative subseries drawn from the training set, which uses randomly-selected shapelets with various dilations, and a ridge regression classifier \citep{guillaume_etal_2022}.

{\multirocket}+{\hydra} combines features from both {\multirocket} and {\hydra}.  {\multirocket} is an extension of {\rocket} and {\minirocket}.  {\rocket} transforms input time series using a large set of random convolutional kernels (random in terms of their length, weights, bias, dilation, and padding), and uses both PPV (`proportion of positive values') and max pooling \citep{dempster_etal_2020}.  {\minirocket} uses a small, fixed set of convolutional kernels and PPV pooling, allowing for highly-optimised computation, and is significantly faster than {\rocket} \citep{dempster_etal_2021}.  {\multirocket} combines the kernels from {\minirocket} with an expanded set of pooling functions, and is close to the most accurate method for time series classification on the datasets in the UCR archive, while being only marginally slower than {\minirocket} \citep{tan_etal_2022}.  {\hydra} combines aspects of both {\rocket} and dictionary methods, counting the occurrence of random patterns, represented by random convolutional kernels, in input time series \citep{dempster_etal_2023}.  {\hydra} is both faster and, with the exception of WEASEL-D, more accurate than other dictionary methods.  All four methods employ a ridge regression classifier by default.

HC2 is an ensemble combining TDE \citep{middlehurst_etal_2021a}, a dictionary method predating WEASEL-D, DrCIF, STC, a shapelet method predating RDST, and Arsenal, an ensemble of {\rocket} models \citep{middlehurst_etal_2021b}.  HC2 is the most accurate method for time series classification on the datasets in the UCR archive.

While there has been significant progress in terms of both accuracy and computational cost since \citet{bagnall_etal_2017}, there is still great variability in the computational efficiency of the most accurate methods, with total compute time on the expanded set of 142 datasets ranging between hours, for the faster methods, and several weeks \citep{middlehurst_etal_2023}.

\section{Method} \label{section-method}

{\quant} involves computing quantiles over a fixed set of intervals on the input time series (and three transformations of the input time series), and using the computed quantiles to train a classifier.  Compared to both DrCIF and rSTSF, we use: (a) a single type of feature (quantiles); and (b) fixed, dyadic intervals.  In contrast to rSTSF, we use no explicit interval or feature selection process (in this sense, feature selection is delegated entirely to the classifier) and, in contrast to DrCIF, we use a standard classifier.  The simplicity of our approach allows for exceptional computational efficiency, and helps to clarify the factors which are material to classification accuracy.

The key characteristics of {\quant} are:

\begin{itemize}
    \item the set of input representations;
    \item the set of intervals;
    \item the features (quantiles); and
    \item the classifier.
\end{itemize}

We implement {\quant} in Python, using the implementation of extremely randomised trees from scikit-learn \citep{pedregosa_etal_2011}.  Our code and results will be made available at \url{https://github.com/angus924/quant}.

\subsection{Input Representations} \label{sec-method-rep}

Following \citet{middlehurst_etal_2021b}, we use the original time series, the first difference, $X'=\{x_{1}-x_{0}, x_{2}-x_{1},\ldots,x_{n-1}-x_{n-2}\}$, and the discrete Fourier transform, $\mathcal{F}(X)$.  We find that it is also beneficial to use the second difference, $X''=\{x'_{1}-x'_{0}, x'_{2}-x'_{1},\ldots,x'_{n-1}-x'_{n-2}\}$, although the improvement in accuracy is marginal: see Section \ref{subsection-sens-rep}.  We find that it is beneficial to smooth the first difference by applying a simple moving average.  Again, the effect seems to be relatively small.  We found no consistent improvement in accuracy by smoothing the other input representations.

\subsection{Intervals} \label{section-method-intervals}

\begin{sloppypar}
    Formally, a time series is a sequence of values ordered in time, ${X = \{x_{0}, x_{1}, \ldots, x_{n - 1}\}}$, where $n$ is time series length.  An interval is a contiguous subset of values $\{x_{a}, \ldots, x_{b}\}$, where $a \geq 0, b > a, b \leq n - 1$.  We define interval length as $m = b - a$, and the number of quantiles per interval as a fraction of interval length, e.g., $m \, / \, 2$ corresponds to computing a number of quantiles equal to half the number of values in a given interval.  For present purposes, we assume that all time series are univariate and of the same length.  We leave the extension of the method to variable-length and multivariate time series to future work.
\end{sloppypar}

\begin{figure}[t]
    \centering
    \includegraphics[width=0.65\linewidth]{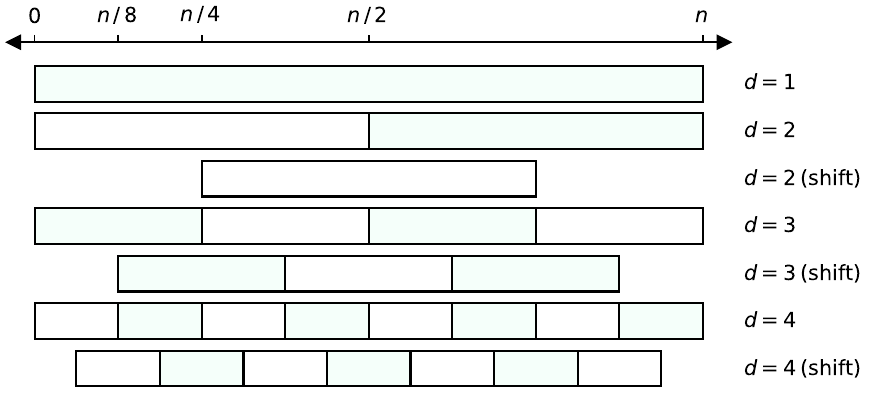}%
    \vspace{-1ex}%
    \caption{An illustration of the set of intervals for a depth of $d=4$, including `shifted' intervals for $d>1$.}%
    \label{fig-dyadic}%
\end{figure}

In contrast to \citet{cabello_etal_2021}, and \citet{middlehurst_etal_2021b}, we use fixed, dyadic intervals.  We define our set of intervals in terms of `depth',~$d$, such that we divide the input time series into $\{2^{0},2^{1},\ldots,2^{d-1}\}$ intervals of length $\{n \, / \, 2^{0}, n \, / \, 2^{1},\ldots, n / 2^{d-1}\}$, as shown in Figure \ref{fig-dyadic}.  For each depth greater than one we also add the same set of intervals shifted by half the interval length.

\pagebreak

Accordingly, the total number of intervals is $2^{d - 1} \times 4 - 2 - d$ for each input representation.  By default, we use a depth of $d=\text{min}(6, \lfloor \log_{2} n \rfloor + 1)$, meaning that there are $120$ intervals per representation, and the smallest intervals are of length $\text{max}(1, n \, / \, 32)$.

As we use nonoverlapping intervals (treating the `shifted' intervals as a separate set of intervals at each depth), the total number of features per depth is always proportional to time series length, $n$, regardless of the number of intervals being constructed.  For example, if we take $m \, / \, 2$ quantiles per interval, for a depth of $d=1$ we take $n \, / \, 2$ quantiles (where $d = 1$, $m = n$), and for a depth of $d = 2$, we likewise take $n \, / \, 2$ quantiles ($m = n \, / \, 2$, so that $2 \times m \, / \, 2 = m = n \, / \, 2$).  Taking $m$ quantiles per interval is equivalent to using the sorted values.

\subsection{Features}

The sorted values represent the empirical distribution of values in each interval.  Quantiles, being a subsample of the sorted values, represent an approximation of the full set of values, that is, an approximation of the empirical distribution.  Importantly, this approximation reduces the size of the feature space, in turn reducing computational complexity (in particular, in relation to classifier training).

As noted above, we define the number of quantiles per interval in proportion to interval length.  By default, we compute $m \, / \, 4$ quantiles per interval, where $m$ is interval length.  (For intervals of length one, we simply use the given value.  For a single quantile, we use the median.)

Broadly speaking, we find that accuracy increases as the number of quantiles per interval increases, although the actual differences in accuracy are small, and computing more quantiles per interval results in proportionally higher computational cost: see Section \ref{subsection-sensitivity-numfeatures}.

\begin{figure}[t]
    \centering
    \includegraphics[width=0.65\linewidth]{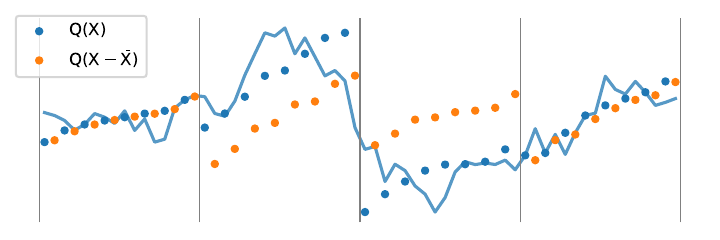}%
    \vspace{-1ex}%
    \caption{An illustration of quantiles drawn from intervals of length $n \, / \, 4$.  We compute quantiles for both the values in each interval, and the values after subtracting the interval mean, representing the distribution of the values and the distribution of the values relative to the mean respectively.}%
    \label{fig-method-quantiles}%
\end{figure}

We find that it is beneficial to compute both: (a) the quantiles of the values in each interval, representing the empirical distribution of the values in each interval; and (b) the quantiles of the values in each interval after subtracting the interval mean, representing the distribution of the values relative to the mean (i.e., such that the values are invariant to level shifts): see Figure \ref{fig-method-quantiles}.  We find that an efficient means of doing this is to alternate between both representations by subtracting the interval mean from every second quantile.  (We only subtract the mean where $m \geq 2$, and the number of quantiles is greater than one.)  Note, however, that the effect of using both the original and mean-corrected quantiles, versus only the original quantiles, or only mean-corrected quantiles, is relatively small: see Section \ref{subsection-sensitivity-mean}.

\subsection{Classifier}

We use extremely randomised trees \citep{geurts_etal_2006}, as per rSTSF \citep{cabello_etal_2021}.  The key distinctions with random forests are that extremely randomised trees do not use bagging, and extremely randomised trees consider a random split point for each candidate feature.

Interval methods can potentially produce a large number of features, depending on the number of input representations, the number of intervals, and the number of features per interval.  For extremely randomised trees, the typical `default' number of candidate features per split is the square root of the total number of features \citep{geurts_etal_2006}.

However, a large number of features in combination with a sublinear number of candidate features per split could potentially result in the classifier `running out' of training examples before adequately exploring the feature space, especially in the context of smaller datasets.  In other words, with a sublinear number of candidate features per split, as the size of the feature space grows, the probability of any given feature being considered decreases.

To this end, we find that it is beneficial to increase the number of candidate features per split to a linear proportion of the total number of features, in particular, $10\%$ of the total number features ($10\%$ of all features are considered at each split).  In effect, we delegate interval and feature selection entirely to the classifier.  The results show that this approach is both effective and computationally efficient.

\pagebreak

\subsection{Complexity}

We treat the computational cost of sorting the values as an upper bound on the cost of computing the quantiles: $O(n \log n)$, where $n$ is time series length.  Naively, computing the quantiles for all intervals requires sorting the values in each interval, for each input representation.  However, as we use a fixed number of input representations, and a fixed number of intervals, we treat these as constant factors.

In principle, we could sort each input representation once, keeping track of the indices of the sorted values, and then form any interval by selecting the already-sorted values using their indices.  In practice, even the naive approach incurs negligible overall computational cost.  Median transform time over 142 datasets in the expanded UCR archive is less than one second.  The majority of compute time is spent in training the classifier.  In other words, any attempt at optimising total compute time should concentrate on reducing the size of the feature space, and/or improving the efficiency of classifier training.  We leave further optimisation for future work.

Assuming approximately balanced trees, the complexity of training the classifier is ${O(p \cdot q \log q)}$, where $p$ is the total number of features, and $q$ is the number of training examples \citep[see][]{louppe_2014}.  As we consider a linear proportion of the total number of features at each split, complexity is linear with $p$ (which is, in turn, linear with $n$).  The number of trees is not proportional to the number of training examples, or the number of features, so we treat this as a constant factor.

\section{Experiments} \label{section-experiments}

We evaluate {\quant} on the datasets in the UCR archive, including 30~datasets recently incorporated into the archive, showing that {\quant} is at least as accurate, on average, as the most accurate existing interval methods, while being meaningfully faster.  We also show the effect of key hyperparameter choices including the number of features, the set of input representations, the number of trees, and the number of candidate features per split.

\subsection{UCR Archive}

We evaluate {\quant} on the datasets in the UCR archive \citep{dau_etal_2019}.  We compare {\quant} with the most accurate existing interval methods, and other state-of-the-art methods for time series classification.  For direct compatibility with published results, we evaluate {\quant} on the same 30 resamples per \citet{middlehurst_etal_2023}.  Results for other methods are taken from \citet{middlehurst_etal_2023}.

\begin{figure}[t]
    \centering
    \includegraphics[width=0.85\linewidth]{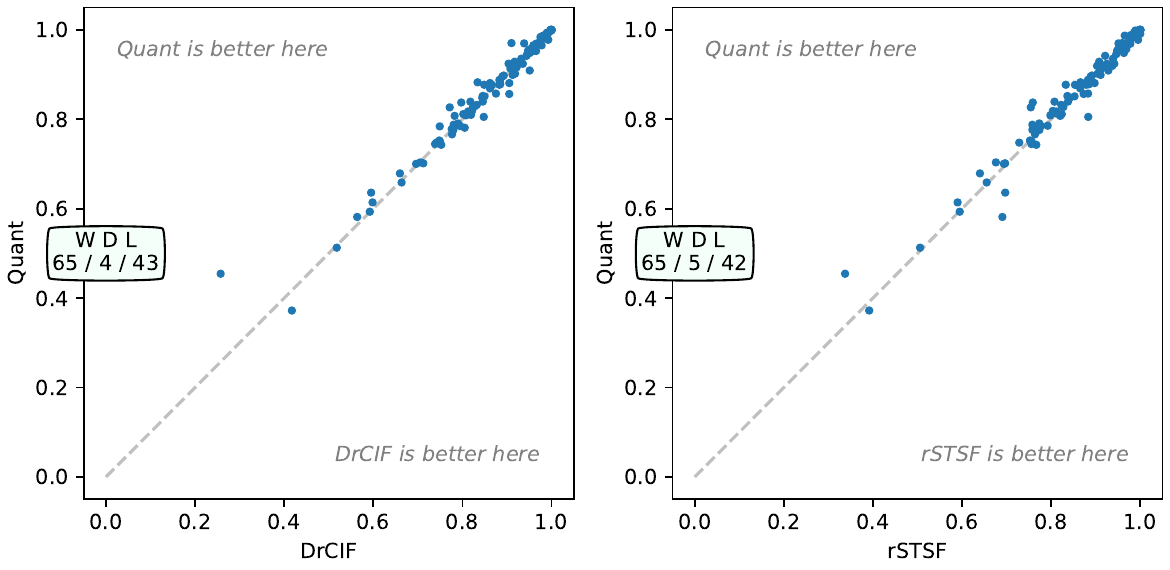}%
    \vspace{-1ex}%
    \caption{Pairwise accuracy of {\quant} vs DrCIF (left), and rSTSF (right), for a subset of 112 datasets from the UCR archive.}%
    \label{fig-pair-ucr112-drcif-rstsf}%
\end{figure}

The difference in mean accuracy, the pairwise win/draw/loss, and the $p$ value for a Wilcoxon signed rank test---between {\quant} and other prominent interval methods, namely, TSF, STSF, rSTSF, CIF, and DrCIF, over a subset of 112 datasets from the UCR archive---are shown in the Multiple Comparison Matrix (MCM) in Figure \ref{fig-mcm-ucr112} on page \pageref{fig-mcm-ucr112}.  In addition, Figure \ref{fig-pair-ucr112-drcif-rstsf} shows the pairwise accuracy of {\quant} versus the two most accurate existing interval methods, namely, DrCIF (left), and rSTSF (right), on the same subset of 112 datasets.

{\quant} is more accurate on average than existing interval methods, including DrCIF and rSTSF, although the actual differences in accuracy are small.  {\quant} is more accurate than DrCIF on 65 datasets, and less accurate on 43.  Similarly, {\quant} is more accurate than rSTSF on 65 datasets, and less accurate on 42.  However, as the results for all three methods are highly correlated, and the differences in accuracy are mostly small, even small changes in accuracy could change the appearance of the results, in particular, the ratio of wins and losses.

\begin{figure}[t]
    \centering
    \includegraphics[width=\linewidth]{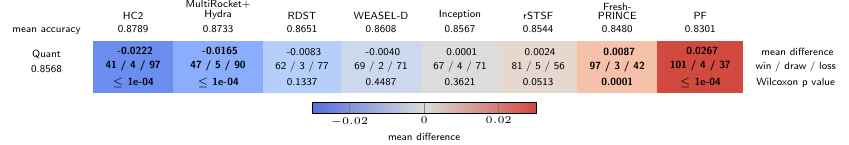}%
    \vspace{-1em}%
    \caption{MCM for {\quant} vs other state-of-the-art methods for 142 datasets from the UCR archive.}%
    \label{fig-mcm-ucr142}%
    \vspace{1em}%
    \bigskip
    \includegraphics[width=0.85\linewidth]{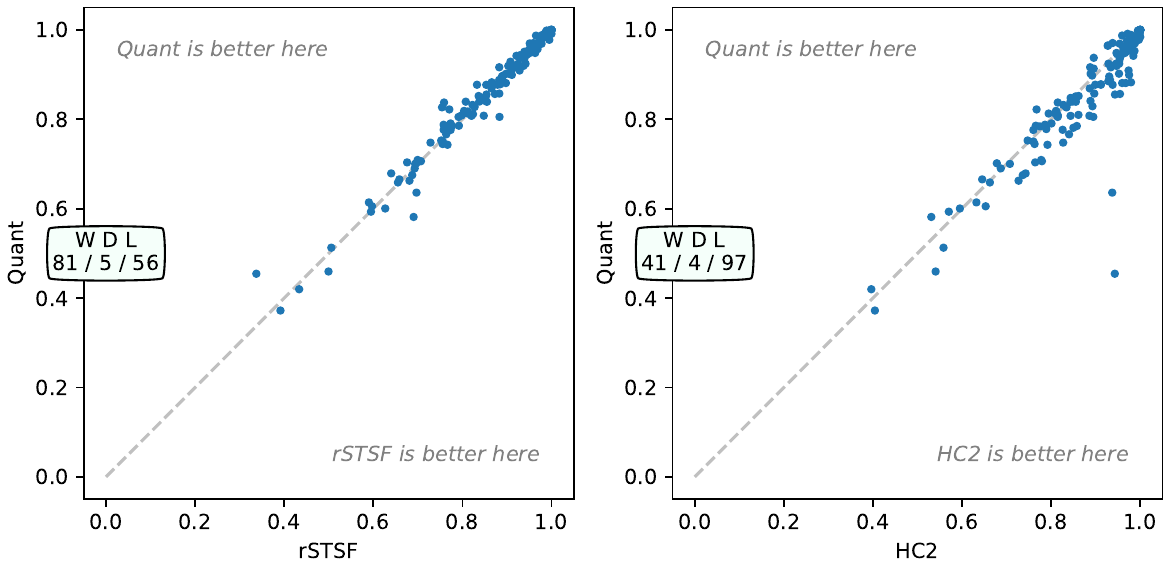}%
    \vspace{-1em}%
    \caption{Pairwise accuracy for {\quant} vs rSTSF (left), and HC2 (right), for 142 datasets from the UCR archive.}%
    \label{fig-pair-ucr142-rstsf-hc2}%
    \vspace{1em}%
\end{figure}

As noted above, thirty additional datasets were added to the UCR archive per the recent `bakeoff redux' \citep{middlehurst_etal_2023}.  Figure \ref{fig-mcm-ucr142} shows the MCM for {\quant} versus current state-of-the-art methods, namely, HC2, {\multirocket}+{\hydra}, RDST, WEASEL-D, InceptionTime, rSTSF, FreshPRINCE, and PF (see Section \ref{section-background}), over 30 resamples of the expanded set of 142 datasets.  Figure~\ref{fig-pair-ucr142-rstsf-hc2} shows the pairwise accuracy of {\quant} versus rSTSF (left), and HC2 (right), for all 142 datasets.

\pagebreak

Over these 142 datasets, {\quant} is reasonably similar to both WEASEL-D and InceptionTime in terms of mean accuracy and win/draw/loss.  However, {\quant} is clearly somewhat less accurate than the most accurate methods (RDST, {\multirocket}+{\hydra}, and HC2).  {\quant} is more accurate than rSTSF on 81 datasets, and less accurate on 56.  In contrast, {\quant} is more accurate than HC2 on only 41 datasets, and less accurate on 97.

However, {\quant} is noticeably faster than any of these methods.  Total compute time (training and inference) over all 142 datasets, averaged over 30 resamples, is less than 15 minutes using a single CPU core, compared to approximately 1 hour 15 minutes for {\multirocket}+{\hydra}, 1 hour 35 minutes for rSTSF, almost 2 hours for WEASEL-D, more than 4 hours for RDST, more than one day for FreshPRINCE, several days for InceptionTime, and several weeks for HC2 and PF.  (Timings for {\quant} are averages over 30 resamples, run on a cluster using Intel Xeon E5-2680 and Xeon Gold 6150 CPUs, restricted to a single CPU core per dataset per resample.  Timings for other methods are taken from \citet{middlehurst_etal_2023}.  Different timings are not necessarily directly comparable, due to hardware and software differences.)  Using 8 CPU cores, compute time is reduced to 6 minutes.

The training time for the classifier is proportional to the total number of features.  Accordingly, we can improve overall compute time by reducing the number of intervals and the number of quantiles per interval.  To this end, Figure \ref{fig-pair-ucr142-fast-rstsf} (Appendix) shows the pairwise accuracy for a faster configuration of {\quant} (informally, {\quant}\textsuperscript{FAST}), using approximately half the number of intervals ($d = 5$), and half the number of quantiles per interval ($m \, / \, 8$), versus rSTSF.  Over 142 datasets, {\quant}\textsuperscript{FAST} is more accurate than rSTSF on 70 datasets, and less accurate on 66.  Total compute time for {\quant}\textsuperscript{FAST} is approximately 7 minutes 40 seconds using a single CPU core.  In other words, {\quant}\textsuperscript{FAST} achieves almost the same accuracy, on average, as rSTSF, but is approximately $10\times$ faster.

\subsection{Sensitivity Analysis}

We demonstrate the effect of key hyperparameters, namely:
\begin{itemize}
    \item the number of features;
    \item the set of input representations (including smoothing);
    \item subtracting the mean; and
    \item the number of trees and the number of features per split.
\end{itemize}

Following \citet{herrmann_etal_2023}, in an effort to avoid the peculiarities of the smallest datasets and the original training/test splits, we conduct the sensitivity analysis using a random sample of 50 of the datasets from the subset of 112 datasets from the UCR archive used in, e.g., \citet{middlehurst_etal_2021b}, using stratified 5-fold cross-validation (such that, for each fold, $80\%$ of the data is used for training, and $20\%$ of the data is used for validation).  In particular, from the subset of 112 datasets, we randomly sample 50 of the 100 datasets where there are at least 100 training examples on an 80/20 split, and at least 5 examples of each class.

\subsubsection{Number of Features} \label{subsection-sensitivity-numfeatures}

\begin{figure}[!t]
    \centering
    \includegraphics[width=0.85\linewidth]{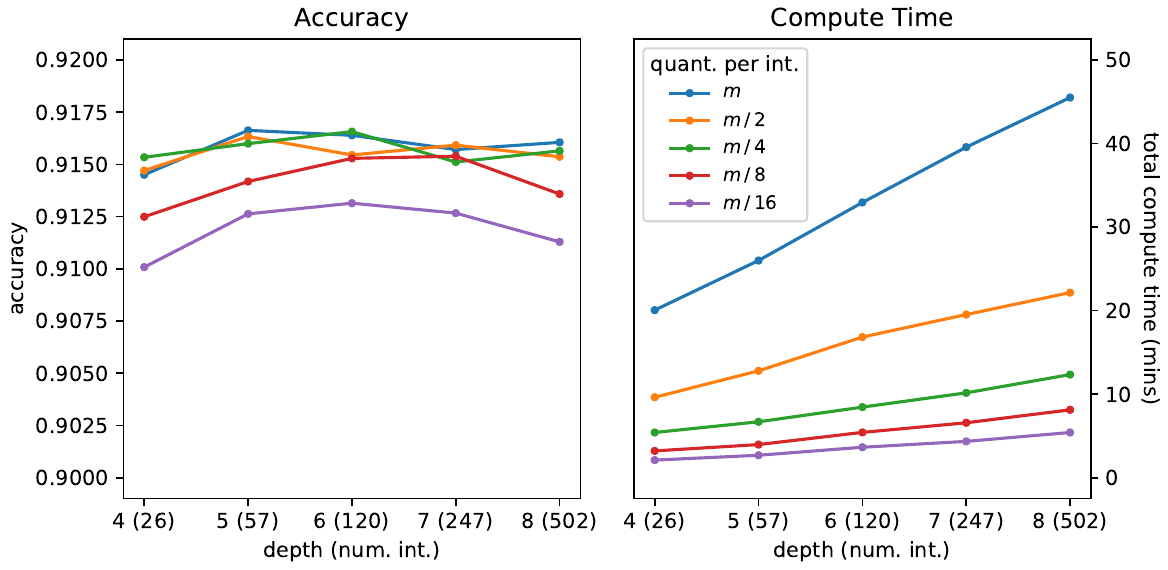}%
    \vspace{-1em}%
    \caption{Mean accuracy (left), and total compute time (right), vs the number of intervals and the number of quantiles per interval.}%
    \label{fig-sens-over-numfeatures}%
    \vspace{1em}%
\end{figure}

Figure \ref{fig-sens-over-numfeatures} shows mean accuracy (left), and total compute time (right), in terms of both: (a) the number of intervals, expressed in terms of depth, $d$; and (b)~the number of quantiles per interval, expressed as a proportion of interval length, $m$.

Accuracy improves modestly as the number of quantiles per interval increases, although the accuracy for $m \, / \, 4$, $m \, / \, 2$, and $m$ quantiles per interval are very similar.  The spread of accuracy values is very small.  However, computing more quantiles per interval results in proportionally greater computational cost due to the expanded feature space: $m$ quantiles per interval requires twice the total compute time of $m \, / \, 2$ quantiles per interval.

It is apparent that, when computing a relatively small number of quantiles per interval, accuracy tends to increase as depth increases, up to a depth of approximately $d=6$, and then decreases.  The same effect is not evident for $m \, / \, 4$ or more quantiles per interval.  We believe that this relates to the balance between distributional information and location information in larger versus smaller intervals: see Section \ref{section-introduction}.  The results suggest that, broadly speaking, larger intervals are more informative than smaller intervals.  With fewer quantiles per interval, more of the information in larger intervals is discarded, and smaller intervals dominate, which leads to lower accuracy.  (It may be possible to counteract this effect by sampling features from larger intervals with higher probability when training the classifier.  We leave this for future work.)  Configurations using more quantiles per interval appear to be relatively immune to this effect.

While increasing depth significantly increases the number of intervals, the corresponding computational cost is linear with depth, as the total number of features computed at each depth is proportional to input length, rather than the number of intervals: see Section \ref{section-method-intervals}.

\begin{figure}[!t]
    \centering
    \includegraphics[width=0.85\linewidth]{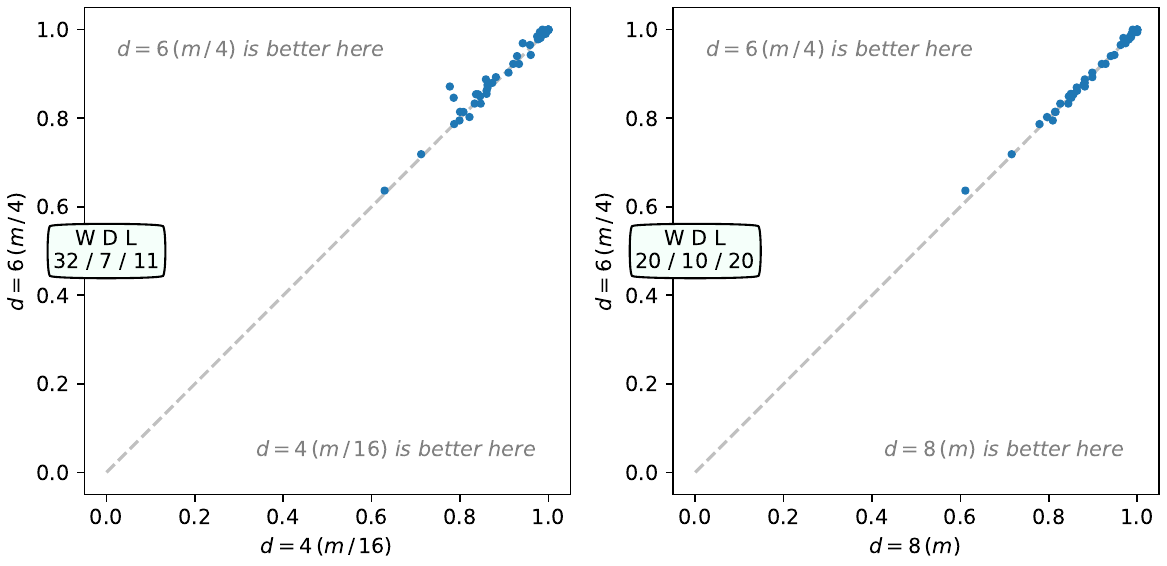}%
    \vspace{-1em}%
    \caption{Pairwise accuracy for a depth of $d=6$ with $m \, / \, 4$ quantiles per interval (the default) vs a depth of $d=4$ with $m \, / \, 16$ quantiles per interval~(left), and a depth of $d=8$ with $m$ quantiles per interval~(right).}%
    \label{fig-sens-pair-numfeatures}%
    \vspace{1em}%
\end{figure}

Figure \ref{fig-sens-pair-numfeatures} shows the pairwise accuracy for a depth of $d=6$ with $m \, / \, 4$ quantiles per interval (the default) versus two extremes in terms of the total number of features, namely, a depth of $d=4$ with $m \, / \, 16$ quantiles per interval (left), and a depth of $d=8$ with $m$ quantiles per interval (right).  While a smaller number of features clearly results in lower accuracy on several datasets, the differences in accuracy compared to a larger number of features are relatively small.

Figure \ref{fig-sens-time-numfeatures} (Appendix) shows compute time versus the number of quantiles per interval for a depth of $d=6$.  This emphasises the extent to which compute time is dominated by the time required to train the classifier which, in turn, is determined by the size of the feature space.

We note that the results presented here relate to the characteristics of the datasets used in these experiments.  In particular, we note that the lengths of most of the time series are relatively short: see Figure \ref{fig-sens-dist-lengths} (Appendix).  In practice, it may be appropriate to adjust the parameters of the transform, e.g., depth, in order to suit the characteristics of a particular dataset.

\subsubsection{Input Representations} \label{subsection-sens-rep}

\begin{figure}[!t]
    \centering
    \includegraphics[width=0.85\linewidth]{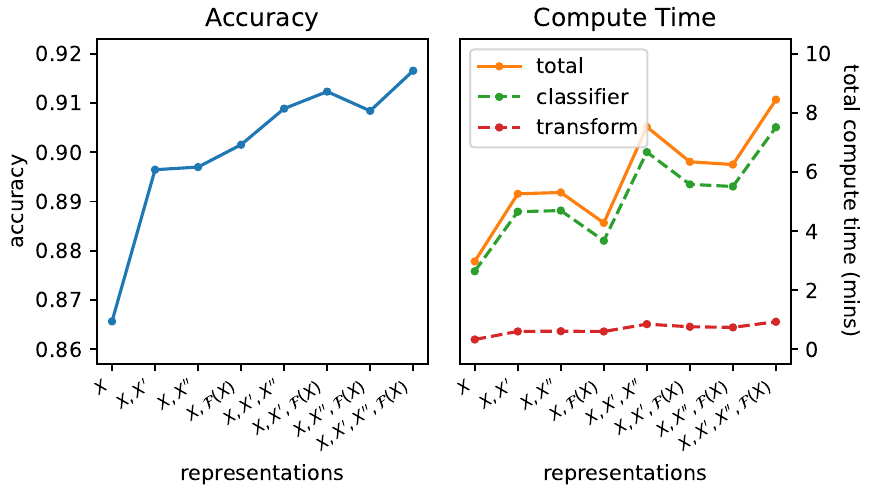}%
    \vspace{-1em}%
    \caption{Mean accuracy (left), and total compute time (right), for different combinations of input representation.}%
    \label{fig-sens-over-rep}%
    \vspace{1em}%
    \bigskip
    \includegraphics[width=\linewidth]{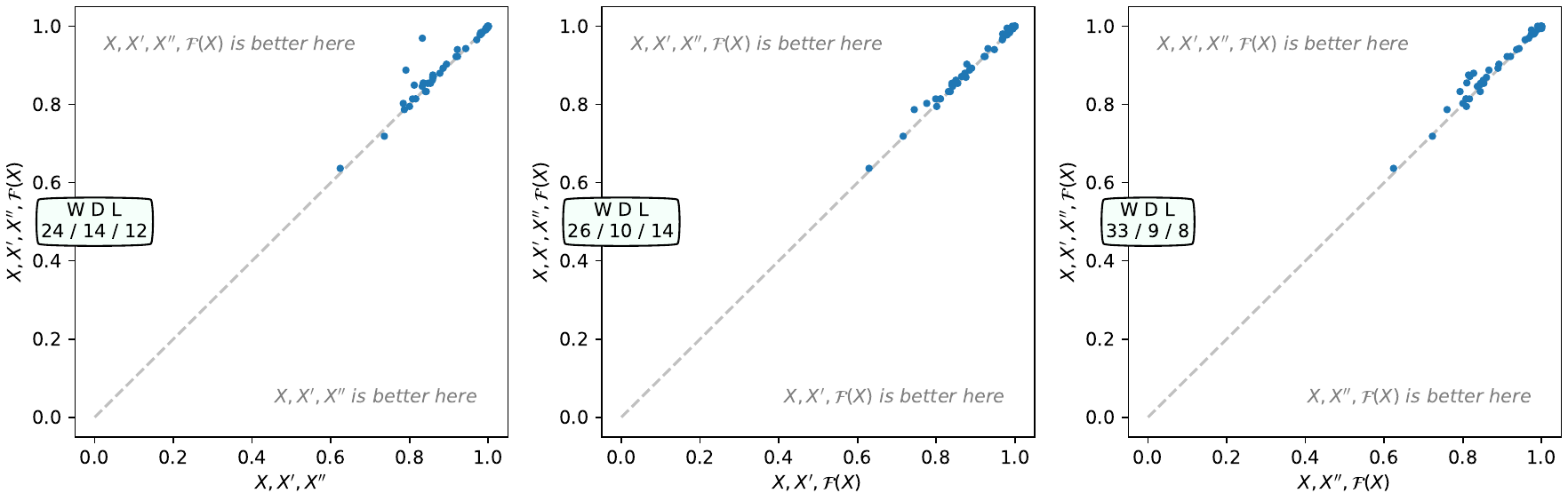}%
    \vspace{-1em}%
    \caption{Pairwise accuracy for all representations (the default) vs removing $\mathcal{F}(X)$ (left), removing $X''$ (centre), and removing $X'$ (right).}%
    \label{fig-sens-pair-rep}%
    \vspace{1em}%
\end{figure}

Figure \ref{fig-sens-over-rep} shows mean accuracy (left), and total compute time (right), for different combinations of input representation.  Figure \ref{fig-sens-pair-rep} shows pairwise accuracy for the default combination of the input time series, $X$, first difference, $X'$, second difference, $X''$, and discrete Fourier transform, $\mathcal{F}(X)$, versus:
\begin{itemize}
    \item $X,X',X''$ (left);
    \item $X,X',\mathcal{F}(X)$ (centre); and
    \item $X,X'',\mathcal{F}(X)$ (right).
\end{itemize}

There is at least some advantage to using each of the three additional representations.  Adding the discrete Fourier transform corresponds to the largest improvements in accuracy on individual datasets (Figure \ref{fig-sens-pair-rep}, left), while adding the second difference makes the least difference (Figure \ref{fig-sens-pair-rep}, centre).  (For all configurations, we maintain the same number of features per representation.)

Figure \ref{fig-sens-pair-smoothing} (Appendix) shows the pairwise accuracy for smoothing versus not smoothing the first difference, via a simple moving average with a window length of $5$.  Smoothing the first difference clearly improves accuracy on several datasets.  We found no consistent improvement in accuracy by smoothing any of the other representations.

\subsubsection{Subtracting the Mean} \label{subsection-sensitivity-mean}

\begin{figure}[!t]
    \centering
    \includegraphics[width=0.85\linewidth]{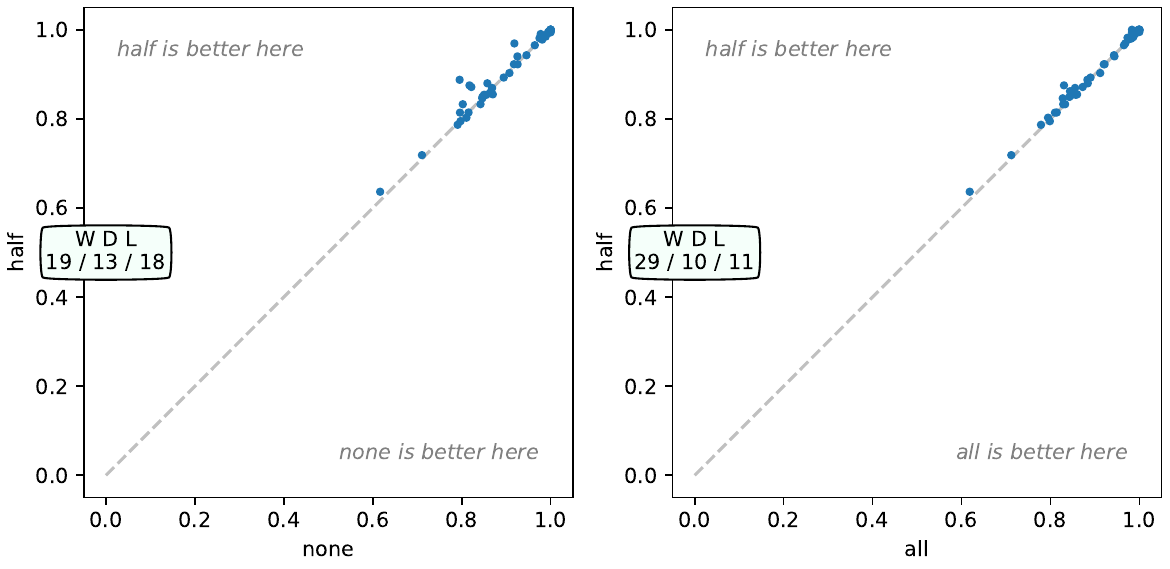}%
    \vspace{-1em}%
    \caption{Pairwise accuracy for subtracting the mean from half of the quantiles (the default) vs not subtracting the mean from any quantiles (left), and subtracting the mean from all quantiles (right).}%
    \label{fig-sens-pair-mean}%
    \vspace{1em}%
\end{figure}

Figure \ref{fig-sens-pair-mean} shows the pairwise accuracy for subtracting the mean from half of the quantiles (the default) versus not subtracting the mean from any quantiles (left), and subtracting the mean from all quantiles (right).  Subtracting the mean from half of the quantiles results in higher accuracy than either not subtracting the mean, or subtracting the mean from all quantiles.  There is no practical effect on compute time.

\subsubsection{Number of Trees}

\begin{figure}[!t]
    \centering
    \includegraphics[width=0.85\linewidth]{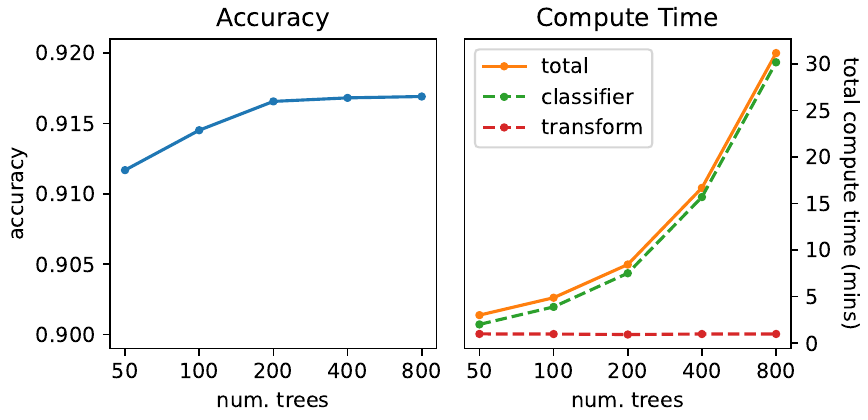}%
    \vspace{-1em}%
    \caption{Mean accuracy (left), and total compute time (right), vs the number of trees.}%
    \label{fig-sens-over-trees}%
    \vspace{0.5em}%
    \bigskip
    \includegraphics[width=0.85\linewidth]{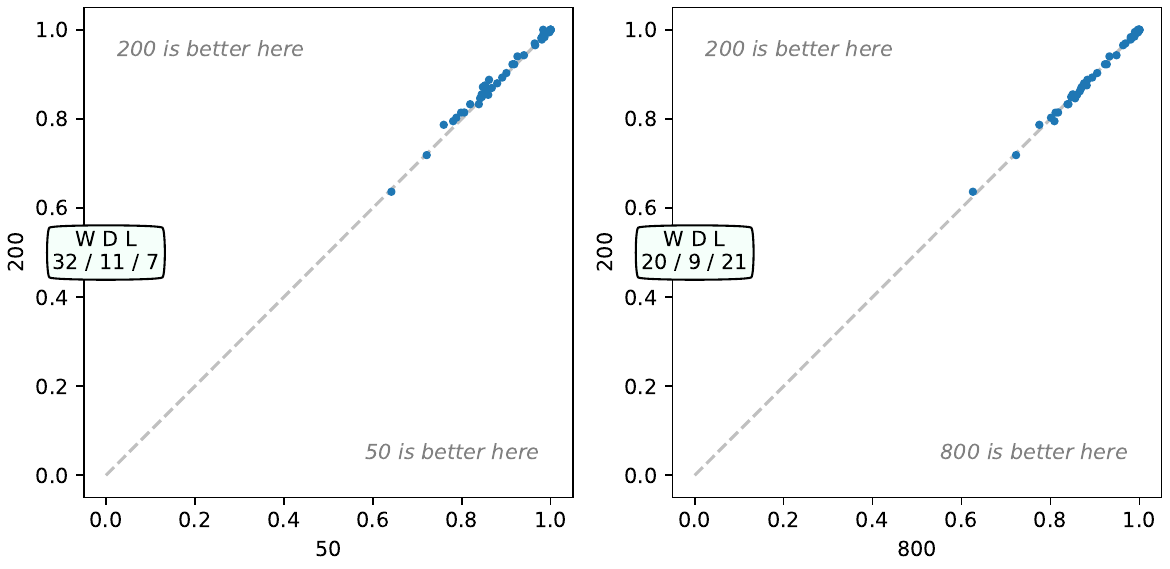}%
    \vspace{-1em}%
    \caption{Pairwise accuracy for 200 trees (the default) vs 50 trees (left), and 800 trees (right).}%
    \label{fig-sens-pair-trees}%
    \vspace{0.5em}%
\end{figure}

Figure \ref{fig-sens-over-trees} shows mean accuracy (left), and total compute time (right), versus the number of trees used in the classifier.  Figure \ref{fig-sens-pair-trees} shows the pairwise accuracy for 200 trees (the default) versus 50 trees (left), and 800 trees (right).

Unsurprisingly, accuracy tends to increase as the number of trees increases, with a proportional increase in computational expense.  However, while there are small but clear differences in accuracy between 50 trees and 200 trees, the differences in accuracy for more than approximately 200 trees are minimal.

\pagebreak

\subsubsection{Number of Features per Split}

\begin{figure}[!t]
    \centering
    \includegraphics[width=0.85\linewidth]{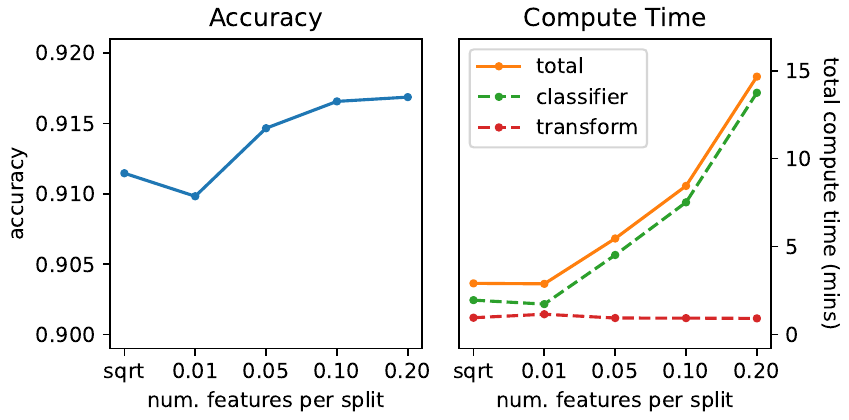}%
    \vspace{-1em}%
    \caption{Mean accuracy (left), and total compute time (right), vs the number of features per split as a proportion of the total number of features.}%
    \label{fig-sens-over-persplit}%
    \vspace{0.25em}%
    \bigskip
    \includegraphics[width=0.85\linewidth]{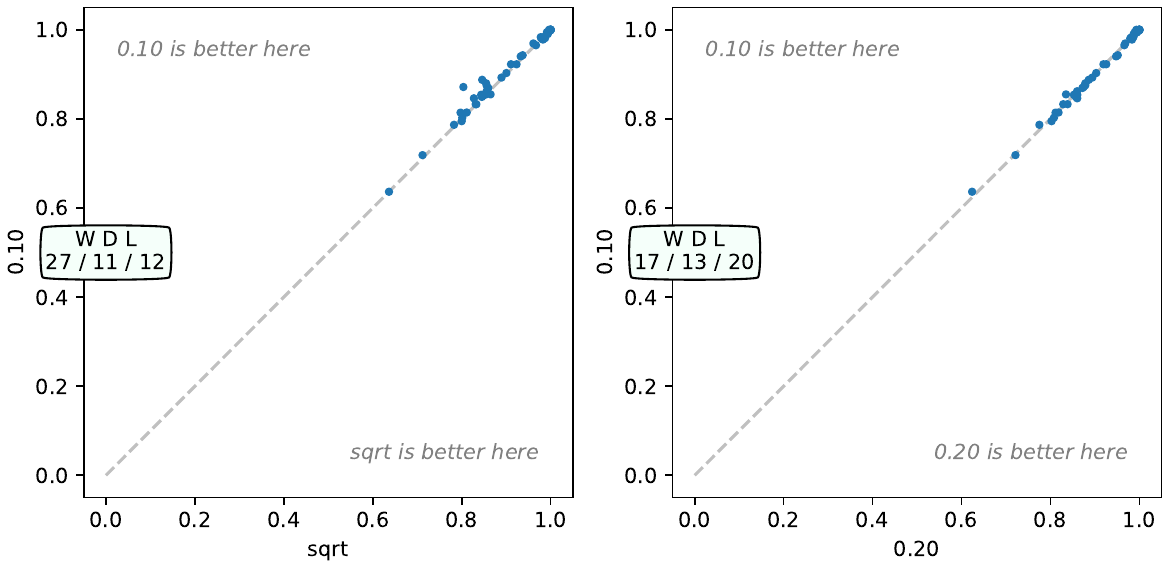}%
    \vspace{-1em}%
    \caption{Pairwise accuracy for $0.1 \times p$ (the default) vs $\sqrt{p}$ (left), and $0.2 \times p$ candidate features per split (right).}%
    \label{fig-sens-pair-persplit}%
\end{figure}

Figure \ref{fig-sens-over-persplit} shows mean accuracy (left), and total compute time (right), versus the number of candidate features per split as a proportion of the total number of features, $p$.  Figure \ref{fig-sens-pair-persplit} shows the pairwise accuracy for $0.1 \times p$ (the default) versus $\sqrt{p}$ (left), and $0.2 \times p$ candidate features per split (right).  Note that $\sqrt{p} > 0.01\times{p}$ for $p < 10{,}000$.

There is a clear advantage in terms of accuracy from increasing the number of candidate features per split to a linear proportion ($\geq 0.05 \times p$) of the total number of features, with a proportional increase in computational expense.  However, the differences in accuracy between sampling $5\%$, $10\%$, or $20\%$ of the features are minimal.

\section{Conclusion}

We demonstrate that a simplified interval method, {\quant}, using a single type of feature (quantiles), fixed intervals, and a standard classifier, without any separate interval or feature selection process, can achieve the same accuracy as the most accurate current interval methods.  Compared to most current state-of-the-art methods for time series classification---many of which require considerable computational resources---{\quant} is both simpler, and represents a significant improvement in terms of accuracy relative to computational cost.  In future work, we intend to explore the extension of the method to variable-length and multivariate time series, as well as further improvements to computational efficiency.

\paragraph*{Acknowledgements}

This work was supported by the Australian Research Council under award DP210100072.  The authors would like to thank Professor Eamonn Keogh and all the people who have contributed to the UCR time series classification archive.

\bibliographystyle{sn-basic}
\bibliography{references}

\clearpage

\appendix

\section*{Appendix: Supplementary Figures}

\null

\vfill

\begin{figure}[h]
    \centering
    \includegraphics[width=0.45\linewidth]{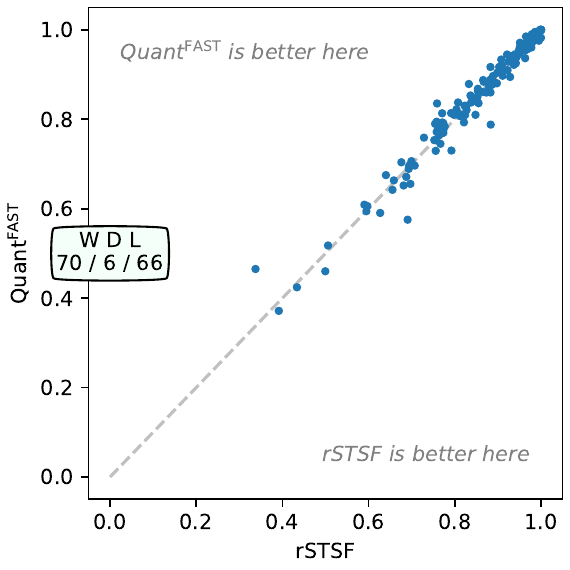}%
    \vspace{-1em}%
    \caption{Pairwise accuracy of {\quant}\textsuperscript{FAST} vs rSTSF for 142 datasets from the UCR archive.}%
    \label{fig-pair-ucr142-fast-rstsf}%
\end{figure}

\vfill

\begin{figure}[h]
    \centering
    \includegraphics[width=0.45\linewidth]{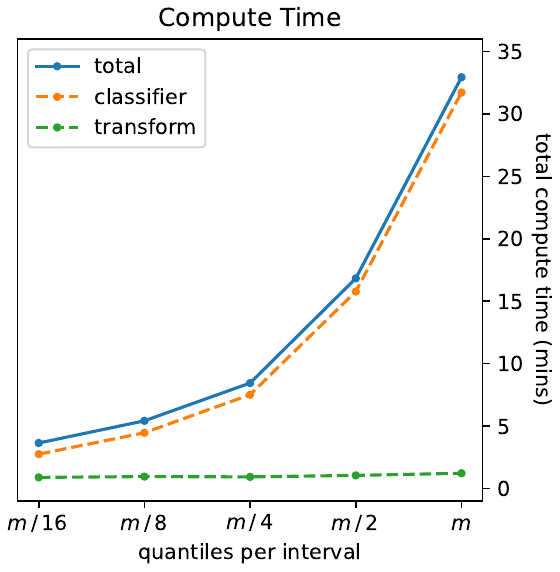}%
    \vspace{-1em}%
    \caption{Compute time (training and inference) vs the number of quantiles per interval for a depth of $d=6$.}%
    \label{fig-sens-time-numfeatures}%
\end{figure}

\vfill

\null

\begin{figure}[h]
    \centering
    \includegraphics[width=0.65\linewidth]{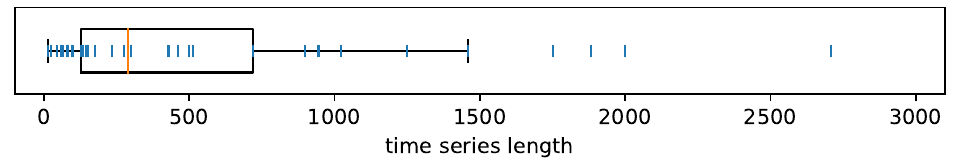}%
    \vspace{-1em}%
    \caption{Distribution of time series length for the datasets used in the sensitivity analysis.}%
    \label{fig-sens-dist-lengths}
\end{figure}

\begin{figure}[h]
    \centering
    \includegraphics[width=0.45\linewidth]{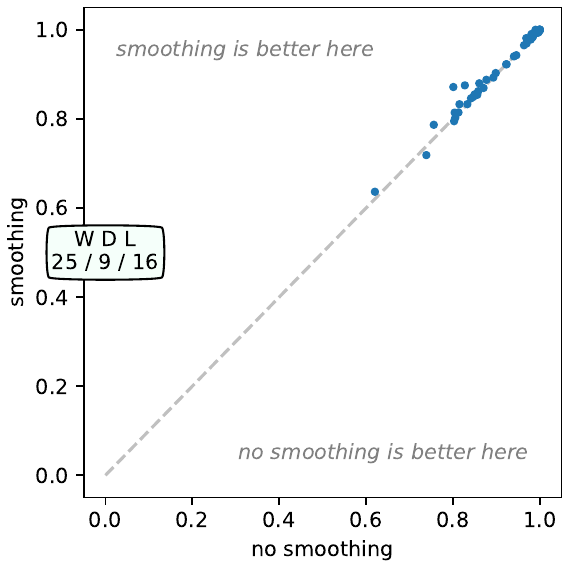}%
    \vspace{-1em}%
    \caption{Pairwise accuracy for smoothing (the default) vs not smoothing the first difference.}%
    \label{fig-sens-pair-smoothing}%
\end{figure}

\end{document}